\documentclass[10pt,twocolumn,letterpaper]{article}

\usepackage{iccv}
\usepackage{times}
\usepackage{epsfig}
\usepackage{graphicx}
\usepackage{amsmath}
\usepackage{amssymb}
\usepackage{bbm}
\usepackage[font=small,labelfont=bf,tableposition=top]{caption}

\usepackage[breaklinks=true,bookmarks=false]{hyperref}

\iccvfinalcopy % *** Uncomment this line for the final submission

\def\iccvPaperID{9516} % *** Enter the ICCV Paper ID here

\ificcvfinal\fi

\usepackage{hyperref}

\title{Contextual Scene Augmentation and Synthesis via GSACNet}

\author{Mohammad Keshavarzi$^{1,2*}$
\hspace{1.5cm} Flaviano Christian Reyes$^{1*}$ \hspace{1.5cm} Ritika Shrivastava$^{1*}$
\\
Oladapo Afolabi$^{1}$ \hspace{1.5cm} Luisa Caldas$^{2}$ \hspace{1.5cm} Allen Y. Yang$^{1}$

\\
 
 \\

$^{1}$FHL Vive Center for Enhanced Reality, University of California, Berkeley
\\
$^{2}$XR Lab, Department of Architecture, University of California, Berkeley

\\

{\tt\small \{mkeshavarzi, fcreyes, ritishri, oafolabi, lcaldas, allenyang\}@berkeley.edu}

}

\begin{document}


\twocolumn[{%
\renewcommand\twocolumn[1][]{#1}%
\maketitle
\ificcvfinal\thispagestyle{empty}\fi
\begin{center}
    \centering
  \includegraphics[width=2\columnwidth]{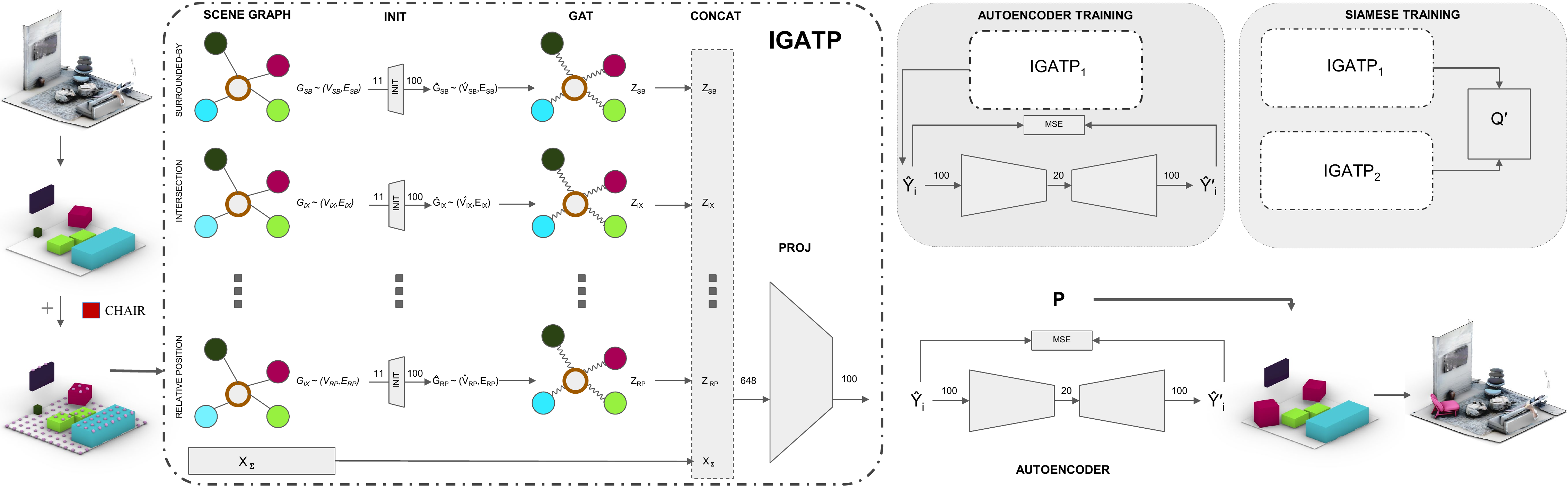}
  \captionof{figure}{To contextually place an object within a scene, GSACNet takes a semantically labeled indoor scene as input and outputs a plausible placement of the object. The system consists of a graph attention, Siamese and auto-encoder network that can be trained with limited scene priors.}~\label{fig:workflow}
\end{center}%
}]

\let\thefootnote\relax\footnotetext{$^*$These authors contributed equally to the work}

%%%%%%%%% ABSTRACT
\begin{abstract}
   Indoor scene augmentation has become an emerging topic in the field of computer vision and graphics with applications in augmented and virtual reality. However, current state-of-the-art systems using deep neural networks require large datasets for training. In this paper we introduce GSACNet, a contextual scene augmentation system that can be trained with limited scene priors. GSACNet utilizes a novel parametric data augmentation method combined with a Graph Attention and Siamese network architecture followed by an Autoencoder network to facilitate training with small datasets. We show the effectiveness of our proposed system by conducting ablation and comparative studies with alternative systems on the Matterport3D dataset. Our results indicate that our scene augmentation outperforms prior art in scene synthesis with limited scene priors available.

\end{abstract}

%%%%%%%%% BODY TEXT

\section{Introduction}
\label{sec:introduction}

Immersive computing via augmented reality (AR) and virtual reality (VR) has demonstrated great potential to become next-generation 3D computing interface. One of the critical bottlenecks in deploying immersive interface in real-world applications is limitations in 3D spatial modeling, namely, many AR/VR experiences are physically constrained by the geometry and semantics of users' local environments where existing furniture and space boundaries are present \cite{keshavarzi2020optimization}. Contrary to placing 2D digital content within a rectangular computer screen, placing virtual 3D asset in real physical space must be consistent with existing furniture and available open spaces. Any immersive experience would become less realistic if asset is placed in locations that are in conflict with existing objects or in conflict with users' common sense. In this paper, we propose to address this problem based on two closely related directions: \emph{contextual scene augmentation} and \emph{contextual scene synthesis}.

First, the challenge of contextual scene augmentation could be mitigated if developers could manually place virtual asset during their design process. However, an end user's local environment is usually not known to the developer at the time of application development, but its model can only be acquired at the moment the application is deployed locally. As a result, it becomes desirable for both the developer and the user to employ fully automatic algorithms to achieve contextual augmentation with the performance as close to the human common sense as possible.

To formulate a good solution for contextual augmentation, it is tempting to adopt the recent trend of using deep neural networks (DNN). Nevertheless, it is also  well known that any modern DNN solution would require large amounts of training data to estimate a set of optimal parameters. This requirement can be met by using either elaborately scanned building datasets \cite{Chang2018, dai2017scannet, song2015sun} or synthetic 3D building datasets \cite{song2017semantic,kolve2017ai2,li2020openrooms,roberts2020hypersim}. In this work, we further propose a novel method to perform critical training data augmentation step in DNN training via contextual synthesis based on real scanned datasets, a good balance between the above two distinct approaches.

Together, our proposed algorithm is called GSACNet, which is an acronym for \emph{Graph Attention SIamese AutoenCoder Network}. Its main contributions are as follows:
\vspace{-2mm}
\begin{enumerate}
    \item GSACNet combines parametric data augmentation techniques with a novel network architecture to achieve plausible indoor scene layouts with small training data. \vspace{-2mm}
    \item By sampling user's target room space, we generate topological scene graphs to represent high-level relationship between objects of the room. This serves as an input to the Graph Attention network followed by a Siamese Network. \vspace{-2mm}
    \item Finally, autoencoder networks cast the plausibility prediction as an anomaly detection problem. Using such workflow we can generate probability maps for an object augmentation in a target scene.
\end{enumerate}

%Our contributions can be summarized follows

%\begin{itemize}
    %\item We introduce a parametric data augmentation workflow for indoor scenes.
    %\item We introduce a novel scene augmentation network architecture by combining a GAT with Siamese and Autoencoder network.
    %\item We show how the combination of our data augmentation method and network architecture can effectively predict contextual augmentations with limited training data. 

%\end{itemize}

\section{Related Work}
\label{sec:review}
\subsection{Scene Synthesis}

Indoor scene synthesis aims to generate a feasible furniture layout of various object classes that satisfy both functional and aesthetic criteria  \cite{zhang2019survey}. Early work of synthetic generation focused on hard-coded rules, guidelines and grammar, following a procedural approach for this problem \cite{bukowski1995object, xu2002constraint, germer2009procedural,merrell2011interactive,Yu2011,yeh2012synthesizing}. The work of \cite{Fisher2012} can be seen as one of the early adapters of example-based scene synthesis using a probabilistic model based on Bayesian networks and Gaussian mixtures.  \cite{savva2017scenesuggest} also proposed a similar approach, involving a Gaussian mixture model and kernel density estimation. In \cite{kermani2016learning}, a full 3D scene was synthesized by adding a single object at a time. This system learned pairwise and higher-order object relations. Work of \cite{Liang2018, Liang2017,fu2017adaptive} also took room functions into account. While object topologies differ in various room functions, a major challenge is that not all spaces can be classified with a certain room function. 
%However, their system targeted an inverse problem of ours, namely, their problem received a selected object location as input and was asked to predict an object type. We find our problem to be more relevant to the needs of a content creator who knows what object they wish to place in scene, but does not have prior knowledge about a user’s surroundings.

More recent works take advantage of deep learning methods, using deep convolutional priors \cite{wang2018deep}, scene-autoencoding \cite{li2019grains}, and new representations of object semantics \cite{balint2019generalized}. The work of \cite{ritchie2019fast} is an example of using deep generative models to sample each object attribute with a single inference
step to allow constrained scene synthesis. This approach was extended in \cite{wang2019planit}, where a combination of object-level and high-level separate convolutional networks were proposed to address constrained scene synthesis problems. Our work comes close to more recent work of \cite{keshavarzi2020scenegen,zhou2019scenegraphnet}, which utilize a scene graph representation to describe a wide variety of object-object and object to room relationships, and tend to conduct constrained scene synthesis by learning from graph priors. Since our work makes use of Graph, Siamese and Autoencoder Networks, we provide more detailed review on these networks in indoor scene synthesis below.

\subsection{Graph Neural Networks}
Graph neural networks have gained immense popularity as a learning methodology for analyzing graphs. Seminal work \cite{monfarGNN} introduced graph neural networks, and the idea of message passing or neighborhood aggregation. On a high level, message passing is an iterative update process used to find node representation by using the graph structure as a means to pass information from neighbors of a target node to the target node itself. Originally, graph neural networks were used as a method to classify nodes within a graph. But over the years, graph neural networks have expanded to autoencode graphs \cite{kipf2016semi}, generate graphs \cite{you2018graphrnn}, solving link prediction problems \cite{zhang2018link}, and segmenting 3D point clouds \cite{mao2019interpolated}. Furthermore, different methods of message passing have been developed as well as their aggregation strategies. Inductive approaches \cite{velivckovic2018deep}, attention mechanisms \cite{velivckovic2017graph}, and gated recurrent units \cite{li2017gated} are some of the more popular approaches. For scene synthesis similar to our scene graph approach, the work of \cite{zhou2019scenegraphnet} utilized a dense scene graph for passing neural messages to augment an input 3D indoor scene with new objects matching their surroundings.

%\begin{figure}
%  \includegraphics[width=1\columnwidth]{figures/Training.pdf}
%  \caption{Training workflow for (left) Siamese and (right) Autoencoder network}~\label{fig:workflow}
%\end{figure}

\subsection{Siamese Networks}
Siamese networks were first introduced in \cite{bromley1994signature} to solve signature verification as an image matching problem. A Siamese neural network consists of twin networks each accepting distinct inputs but joined by an energy function at the top. This function computes some metric between the highest-level feature representation on each side \cite{koch2015siamese}. Siamese networks have been used in various applications of indoor design and floorplanning due to its ability to learn from limited data. For example, \cite{guo2018learning} used Siamese networks for scene change detection. 

\subsection{Autoencoders}
Autoencoders \cite{bourlard1988auto, hinton1994autoencoders} are a type of neural networks designed to map high-dimensional input data to a low-dimensional latent representation that captures most of the important information needed to reconstruct the data back. This is achieved by a sequence of nonlinear mapping (encoder network) from the input space to a latent space followed by another sequence of nonlinear mapping (decoder network) from the latent space back to the input space. The parameters in these mapping networks are chosen to minimize the difference between the input data and its reconstructed image. A byproduct of this design is that autoencoders may be used to detect data that strays from the input distribution during training. The idea is that the low-dimensional latent space is forced to capture only information about the subset of the input space that data is drawn from, and accurately reproduce data living in this space. All other data points will suffer a significant  loss in fidelity when mapped through the autoencoder. Consequently, autoencoders have been used for anomaly detection in a variety of applications \cite{zong2018deep,zhou2017anomaly, sakurada2014anomaly}. Variations of autoencoders, and other similar networks such as Generative-Adverserial Networks have been used as tools for generating plausible indoor scenes by sampling from their associated latent spaces \cite{purkait2020sg, li2019grains}. In this work, we make use of autoencoders to discriminate between natural looking and random scene arrangements, and separate a scene proposal and scene generation as two independent tasks.

\subsection{Data Augmentation}
Data Augmentation is referred to a class of techniques that aim to enhance the size and quality of a given training dataset. Such techniques can result in improving the generalization performance of deep neural networks while avoiding the problem of overfitting when trained on limited data. A wide variety of methods such as geometric transformations, color space augmentations, kernel filters, mixing images, random erasing, feature space augmentation, adversarial training, generative adversarial networks, neural style transfer, and meta-learning have been explored in this field. \cite{shorten2019survey} provides a survey on image data augmentation for deep learning approaches.  Our approach takes advantage of a widely used method in computer-aided design in architecture, commonly known as parametric design \cite{caetano2020computational}. In parametric design, various design elements in a procedural model can be transformed using input parameters while maintaining their topological relationships with each other. Work of \cite{keshavarzi2020genscan, keshavarzi2020sketchopt, keshavarzi2021genfloor} are examples of systems that utilize parametric workflows for generating various space layout configurations.

\section{Methodology}

Figure \ref{fig:workflow} shows the general workflow of our system. Given a semantically segmented target room, our system aims to contextually place objects within the scene while maintaining plausible relationship with the room and its objects. To do so, the room space will be sampled uniformly where the sample points are considered the center of possible placement, and the plausibility probability at each sample is then calculated. The GSACNet architecture involves five modules: (1) scene graph extraction; (2) initialization; (3) graph attention; (4) projection into learned space; and (5) plausibility assessment via an autoencoder network. The integral copies of the first four modules together are called IGATP (Initialization Graph ATtention Projection), and the modules are then used for Siamese training; the autoencoder is trained separately. In the following subsections, we first define the topological relationships in which the scene graphs utilize, followed by the formulations of the various components of our network architecture. 

\subsection{Definitions}
We consider a room or a scene in 3D space where its floor is on the flat $(x,y)$-plane and the $z$-axis is orthogonal to the $(x,y)$-plane. We denote the room space in a floor-plan representation as $R$, namely, an orthographic projection of its 3D geometry plus a possible adjacency relationship that objects in $R$ may overlap on the $(x,y)$-plane but on top of one another along the $z$-axis. This can also be viewed as a 2.5-D representation of the space. 

Further denote the $k$-th object (e.g., a bed or a table) in $R$ as $O_{k}$. The collection of all $n$ objects in $R$ is denoted as $\mathcal{O}= \{O_1, O_2,... O_n\}$.  $B(O_k)$ represents the bounding box of the object $O_k$. $\dot{O}_k$ represents the center of the object $O_k$. For convenience, we will also define $\dot{O_k}_{xy}$ as $\dot{O_k}$ but projected onto either existing furniture below or the floor plane of $R$. Every object $O_k$ has a furniture group to classify its type. The set of all furniture groups is called $G$ = $\{g_1,..., g_m\}$, where each group $g_i$ contains all objects of the same furniture group.
%Furthermore, each $O_k$ has a primary axis $a_k$ and a secondary axis $b_k$. For Frontal Facing objects, $a_k$ represents the orientation of the object. $a_k$ and $b_k$ are both unit vectors such that $b_k$ is a $\frac{\pi}{2}$ radian counter clockwise rotation of $a_k$.  We define $\theta_{a_k}$ and $\theta_{b_k}$ to be the angle in radians represented by $a_k$ and $b_k$ respectively. Additionally, each $O_k$ has a radius vector $r_k = [M_{a_k}, M_{b_k}, M_{a_k \times b_k}]$ which stores the scaling of $B(O_k)$ along of $a_k$, $b_k$, and $a_k \times b_k$. $B(O_k)$ utilizes $M_{a_k}a_k$, $M_{b_k}b_k$, and $M_{a_k \times b_k}(a_k \times b_k)$ to construct the bounding box vertices and thus its 6 planes.
Furthermore, $\dot{O_k}$ is the centroid of $B(O_k)$.

 For each room $R$, we define $\mathcal{W} = \{W_1, W_2, ..., W_l\}$ where each $W_k$ is a wall of the $l$-sided room. In the floor plan representation, $W_k$ is represented by a 1D line segment. We also introduce a distance function $\delta(a, b)$ as the shortest distance between a and b objects. For example, $\delta(B(O_{k}),\dot{R})$ is the shortest distance between the bounding box of $O_k$ and the center of the room $R$. Intersection of bounding boxes is regarded as $\delta(B(O_k), B(O_j)) = 0$.

\subsection{Spatial Relationships}
%This section describes the spatial relationships in a scene that we utilize for scene representation extraction.
%(see section 3.5).
 %We include pairwise relationships between objects (eg. between a chair and a desk) and between object groups (eg. between a dining table and dining chairs). Additionally, we include relationships between an object and the room.

% Each scene, $R$ is  the 2D orthographic projection of a room. 

\subsubsection{Object to Room Relationships}
\paragraph{RoomPosition:} The room position feature of an object denotes whether an object is at the middle, edge, or corner of a room. This is based on how many walls are less than a distance $\rho$ away from an object. For convenience, we define $\phi$ as follows:
\begin{equation}
    \label{eqn:room_position}
    \phi\left(O_k, W_i\right) = \mathbbm{1}(\delta (B(O_{k}), W_i) < \rho).
\end{equation}
Using $\phi$, we define RoomPos as follows:
% \begin{equation}
%     \label{eqn:room_position}
%     \mbox{RoomPos}\left(O_k, R\right) = \sum_{W_i \in \mathcal(W)} \mathbbm{1}(\delta (B(O_{k}), W_i) < \rho).
% \end{equation}
\begin{equation} 
    \label{eqn:room_position} 
        \mbox{RoomPos}\left(O_k, R\right) = \sum_{W_i \in \mathcal(W)} \phi(O_{k}, W_i). 
\end{equation}
In other words, if $\mbox{RoomPos}(O_k, R) \geq 2$, the object is near at least 2 walls of a room, and hence is near a \textit{corner} of the room. If $\mbox{RoomPos}(O_k, R) = 1$, the object is near only one wall of the room and is at the \textit{edge} of the room. Otherwise, the object is not near any wall and is in the \textit{middle} of the room.  % Along the x-axis, $\rho$ is set to 20\% of the width of the $R$, and along the y-axis, $\rho$ is set to 20\% of the length;  
 
\subsubsection{Object to Object Group Relationships}
\paragraph{AverageDistance:} For each object and each group of objects, we calculate the average distance between that object and all objects within that group. 
% \begin{equation}
%     \mbox{AvgDist}(O_k, g_i) = 
%     \dfrac{\sum\limits_{\substack{O_j \in g_i}} \delta(B(O_k), B(O_j))}{ |\{O_j \in g_i\}| }.
% \end{equation}
\begin{equation}
    \mbox{AvgDist}(O_k, g_i) = 
    \dfrac{\sum\limits_{\substack{O_j \in g_i}} \delta(B(O_k), B(O_j))}{ |\{g_i\}| }.
\end{equation}

\paragraph{SurroundedBy:} For each object and each group of objects, we compute how many objects in the group are within a close proximity of an object. Suppose $O_k$ and $O_j$ are within room $R$. $O_j$ is within the proximity of $O_k$ if $\delta(B(O_k), B(O_j)) < \epsilon_k = \|[L_k, W_k]\|_2$, where $L_k, W_k$ refer to the length and width of $B(O_k)$, respectively. For convenience, we define a function $\sigma$ as follows:
\begin{equation}
    \mbox{$\sigma$}\left(O_k, O_j\right) =
    \mathbbm{1}( \delta\left(B(O_k), B(O_j)) < \varepsilon_k \right).
\end{equation}
Using $\sigma$, we define the surrounded-by function SurrBy as follows:
\begin{equation}
    \mbox{SurrBy}\left(O_k, g_i\right) =
    \sum\limits_{\substack{O_j \in g_i}}  \sigma(O_k, O_j).
\end{equation}

\paragraph{IntersectionXY:} For each object and each group of objects, we compute how many objects in the group are intersecting an object in the XY plane. Suppose $O_k$ and $O_j$ are within room $R$. $O_j$ intersects $O_k$ in the XY plane if $\delta(B_{xy}(O_k), B_{xy}(O_j)) = 0$. $B_{xy}(O_k)$ refers to the bounding box of $O_k$ projected onto the ground floor plane of $R$. For convenience, we define a function $\iota$ as follows:
\begin{equation}
    \mbox{$\iota$}\left(O_k, O_j\right) =
    \mathbbm{1}( \delta\left(B_{xy}(O_k), B_{xy}(O_j)) = 0 \right).
\end{equation}
Using $\iota$, we define the intersection-XY function InterXY as follows:
\begin{equation}
    \mbox{InterXY}\left(O_k, g_i\right) =
    \sum\limits_{\substack{O_j \in g_i}}  \iota(O_k, O_j).
\end{equation}

\paragraph{Co-Occurence:} Given a room $R$, an object $O_k$ and another object $O_j, k \neq j$ are said to co-occur if they exist within $R$. 
\begin{equation}
    \mbox{Cooc}\left(R, O_k, O_j\right) = \mathbbm{1}( O_k, O_j \in R \wedge k \neq j).
\end{equation}

\subsubsection{Object Support Relationships} \label{sec:support-definition}
\paragraph{Support:}
An object is considered to be supported by a group if it is on top of an object from the group, or supports a group if it is underneath an object from the group. Due to erroneous bounding box intersections within our dataset, we relax the definition of support by enforcing a threshold $\tau$ on the separation distance between the bottom bounding box plane of the top object and the top bounding box plane of the bottom object. %This threshold is equal to 0.05 meters. 
For convenience, we define a function $\psi$ as follows:
\begin{equation}
    \mbox{$\psi$}\left(O_k, O_j\right) = 
    \begin{cases} 
      1 & \mbox{ $0<B(O_k)_{bottom}$ - $B(O_j)_{top} < \tau$}; \\ 
      -1 & \mbox{ $0<B(O_j)_{bottom}$ - $B(O_k)_{top} < \tau$};  \\
      0 & \text{otherwise}.
   \end{cases}
\end{equation}
% Using $\psi$, we define the support function Supp as follows:
% \begin{equation}
%     \mbox{Supp}\left(O_k, g_i\right) = 
%     \begin{cases} 
%       1 & \exists O_j \in g_i, \mbox{ \psi($O_k$, $O_j$) = 1}; \\ 
%       -1 & \exists O_j \in g_i, \mbox{ \psi($O_k$, $O_j$) = -1};  \\
%       0 & \text{otherwise}.
%    \end{cases}
% \end{equation}
Using $\psi$, we define more specific support relationships. Specifically, the function SuppBy describes the number of objects that support $O_k$:
\begin{equation}
    \mbox{SuppBy}\left(O_k, g_i\right) =
    \sum\limits_{\substack{O_j \in g_i}}  \mathbbm{1}\left( \psi(O_k, O_j) = 1) \right).
\end{equation}
Similarly, the function SuppTo describes the number of objects that $O_k$ is supporting:
\begin{equation}
    \mbox{SuppTo}\left(O_k, g_i\right) =
    \sum\limits_{\substack{O_j \in g_i}}  \mathbbm{1}\left( \psi(O_k, O_j) = -1) \right).
\end{equation}

%\subsection{Extracting Scene Representations}
%This section describes how our system encodes spatial relationship information within a given room $R$ by constructing (1) a summary vector and (2) a set of scene graphs. First, suppose there are $m$ total furniture groups.
%: bed, chair, storage, decor, picture, table, sofa, and TV. These groups serve as the labelling for each furniture group that is present in our dataset. % Secondly, when mentioning a target object, we are referring to the object that we want to place in the room. For convenience, we will refer to the target object at $O_k$. 
%Additionally, suppose that the target object is placed in R such that its ground floor center is $\dot{O_k}_{xy}$, and we extract spatial relationships based on this placement. %Finally, note that distance between objects is measured by $\delta$ using object bounding boxes as input (see section 3.2).

\begin{figure*}
  \centering
  \includegraphics[width=2\columnwidth]{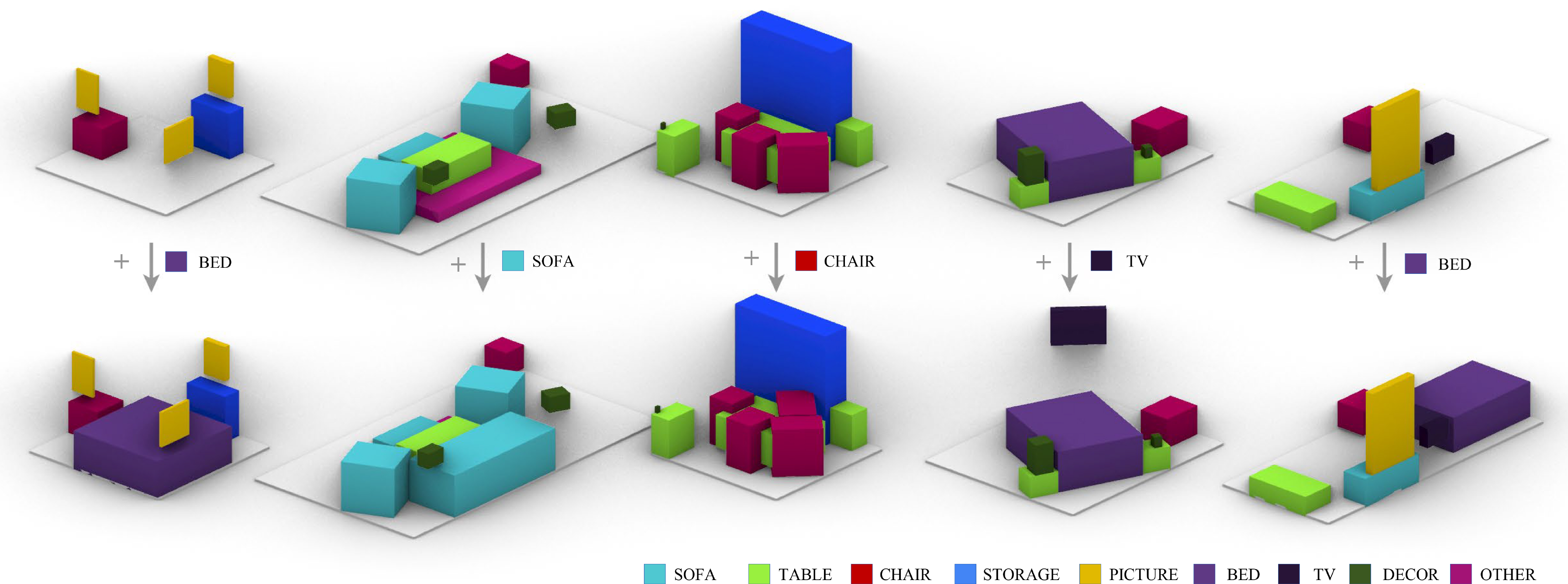}
  \caption{Example of contextual scene augmentation results. Top row illustrates the target scene, and bottom row illustrates the augmented scene. }~\label{fig:sceneAug}
\end{figure*}

\subsection{Network Architecture}\label{sec:netarch}
\subsubsection{Scene Graph Extraction}
 We sample points uniformly in the $(x, y)$-plane. Regarding each point as the center of possible placement for a target object $O_k$ on the ground floor plane (the sample point will be temporarily considered as $\dot{O_k}(x,y)$), we form a summary vector $X_{O_k}$ and scene graphs $\mathcal{G}_{r} \sim (V_{r}, E_{r})$ for each relationship $r \in \mathcal{R}$, where $\mathcal{R}$ is the set of all spatial relationships under consideration. As an important aside, in our scene synthesis system, there is a model per furniture group. For the target object $O_k$, we will use the model associated with its furniture group $g_{O_k}$. We denote such model usage by sub-scripting the main modules by $g_{O_k}$.

We use homogeneous scene graphs to represent spatial relationships. %In particular, if a directed edge from node A to node B exists within a scene graph, then some spatial relationship exists between the furniture object associated with node A and the furniture object associated with node B. 
Objects are defined as nodes, and relationships are defined as edges. The target object node refers to the node in a scene graph associated with the object we want to place in the scene (the target object $O_k$). We refer to this node as $v_{O_k}$. Secondly, given $O_a$ and $O_b$ both in a room $R$, the edge from node $v_{O_a}$ to node $v_{O_b}$ is referred to as $e_{(a, b)}$. In the following paragraphs, we will describe the connection criteria per scene graph that we use in our system. A scene graph's connection criteria refers to the rules that determine whether or not there exists an edge between two nodes. %Our system pays attention to (1) bounding box intersections, (2) objects surrounding other objects, (3) objects being supported by other objects, (4) objects supporting other objects, (5) relative positioning to walls, and (6) co-occurrence with other objects. 
% Our choice of scene graphs is inspired from prior work of \cite{zhou2019scenegraphnet} and \cite{keshavarzi2020scenegen} in addition to new modalities introduced in this paper. 
We utilize homogeneous scene graphs such as in \cite{zhou2019scenegraphnet}, and we utilize spatial relationship criterion from \cite{keshavarzi2020scenegen} to construct those scene graphs. However, we introduce the intersection scene graph as a new spatial relationship consideration. 

% For clarity, the target object node refers to the node in a scene graph associated with the object we want to place in the scene (the target object $O_k$). For convenience, lets call this node $v_{O_k}$. Secondly, given $O_a$ and $O_b$ both in a room $R$, the edge from node $v_{O_a}$ to node $v_{O_b}$ is referred to as $e_{(a, b)}$.
\begin{itemize}
\item{Intersecting Objects Scene Graph.}
% This scene graph captures bounding box intersections that occur within a room $R$, relative to $O_k$. For convenience, we refer to the intersecting objects scene graph as $\mathcal{G}_{IXY} \sim (V_{IXY}, E_{IXY})$. The connection criteria for $\mathcal{G}_{IXY}$ is as follows:
\begin{equation}
    E_{IX} = \{e_{(j, k)}|\forall O_j \in R, j\neq k, \iota(O_k, O_j) = 1\}
\end{equation}
% Another way to think about the connection criterion is that if another object's bounding box is intersecting the target object's bounding box, then an edge is connected from the other object's node to the target object's node.

\item{Surrounded-By Scene Graph.}
% This scene graph captures what objects surround $O_k$. For convenience, we refer to the surrounding objects scene graph as $\mathcal{G}_{SB} \sim (V_{SB}, E_{SB})$. The connection criteria for $\mathcal{G}_{SB}$ is as follows:
\begin{equation}
    E_{SB} = \{e_{(j, k)}|\forall O_j \in R, j\neq k, \sigma(O_k, O_j) = 1\}
\end{equation}
% Another way to think about the connection criterion is that if another object is within the proximity of the target object, then an edge is connected from the other object node to the target object node. 

\item{Support-By Objects Scene Graph.}
% This scene graph is primarily concerned with what objects support the target object. For convenience, we refer to the support-by objects scene graph as $\mathcal{G}_{SBY} \sim (V_{SBY}, E_{SBY})$. The connection criteria for $\mathcal{G}_{SBY}$ is as follows:
\begin{equation}
    E_{SBY} = \{e_{(j, k)}|\forall O_j \in R, j\neq k, \psi(O_k, O_j) = 1\}
\end{equation}
% Another way to think about the connection criterion is if another object is on top of the target object (or within the threshold defined in section 3.3.3), then an edge is connected from the other object to the target object node.

\item{Support-To Objects Scene Graph.}
% The support-to objects scene graph is almost exactly the same as $\mathcal{G}_{SBY}$, but now, we consider the objects that support the target object. For convenience, we refer to the support-to objects scene graph as $\mathcal{G}_{STO} \sim (V_{STO}, E_{STO})$. The connection criteria for $\mathcal{G}_{STO}$ is as follows:
\begin{equation}
    E_{STO} = \{e_{(j, k)}|\forall O_j \in R, j\neq k, \psi(O_k, O_j) = -1\}
\end{equation}

\item{Relative Position Scene Graph.} 
% This scene graph is primarily concerned with where the object is in the room, relative to the walls in the room. The nodes in this graph are associated with the walls, floor, and target object. 
Suppose $W_R$ refers to the set of walls of room $R$ and $F_R$ refers to the floor of $R$. Furthermore, $W_i \in W_R$ is defined similarly as $O_j \in R$, but $W_i$ is signified to represent a wall object. The same is true of $F_R$. Lastly, $e_{(w_j, k)}$ refers to the edge from $v_{W_j}$ and $v_{O_k}$, and $e_{(f, k)}$ refers to the edge from $v_{F_R}$ to $v_{O_k}$. % For convenience, we refer to the relative position scene graph as $\mathcal{G}_{RP} \sim (V_{RP}, E_{RP})$. The connection criteria for $\mathcal{G}_{RP}$ is as follows:
\begin{equation}
    E_{RP,W} = \{e_{(w_j, k)}|\forall W_j \in W_R, \phi(O_k, W_j) = 1\}
\end{equation}
\begin{equation}
    E_{RP,F} = \{e_{(f, k)}|E_{RP,W} = \emptyset\}
\end{equation}
\begin{equation}
    E_{RP} = E_{RP,W} \cup E_{RP,F}
\end{equation}
Another way to describe $E_{RP}$ is if a wall is within the proximity of an object, then an edge is drawn from the node associated with the wall to the target object. If no walls meet this criteria, an edge is drawn from the floor node to the target object node.

\item{Co-occurring Scene Graph.}
% This scene graph keeps track of what objects exist with the target object through edge connections. Each object within the room has an associated node in this graph. An edge is drawn from each of these nodes to the target node. For convenience, we refer to the co-occuring scene graph as $\mathcal{G}_{CO} \sim (V_{CO}, E_{CO})$, and the edge criterion is as follows:
\begin{equation}
    E_{CO} = \{e_{(j, k)}|\forall O_j \in R, \text{Cooc}(R, O_k, O_j) = 1\}
\end{equation}

\item{Graph Feature Vectors and Default Nodes.}
Nodes have a feature vector associated with them. In particular, the node feature vector is in 11-D space, where the first 10-D represent the one-hot encoding of the object furniture group (first 8-D for the furniture groups and the last 2-D are for the walls and floors, respectively) and the last dimension represents the distance ordering from the target node $O_k$. Distance ordering refers to an object's rank of how close they are to the target object. For instance, suppose there is a table, a chair, and a bed in the room. The table is considered to the be target object, and the chair is closer to the table than the bed. Then, the table receives a distance order of 0, the chair receives a distance order of 1, and the bed receives a distance order of 2. 

In each scene graph, there also exists a default node such that its feature vector is a zero vector, except for the component associated with relative ordering. For default nodes, the relative ordering is set to -1. The an edge exists from the default node to the target object, and if only the default node exists within a scene graph, then no objects met the connection criteria for the specific scene graph.
\end{itemize}

\subsubsection{Summary Feature Vector}

For a proposed floor plane centering $\dot{O_k}_{xy}$ of object $O_k$ in room $R$, the summary vector $X_{O_k}$ can be described as follows:
$$ X_{O_k} = \begin{bmatrix} 3C, & EB, & CB, & AD, & SB, \\ 
                                IX, & SBY, & STO \end{bmatrix} \in R^{48} $$

\paragraph{$3_{\text{closest}}$ ($3C$):}

\cite{zhou2019scenegraphnet} utilizes an ordered aggregation scheme for message passing. Specifically, messages are passed through a GRU in the order of farthest object-node to closest object-node. Inspired by this idea, our summary vector takes into account the three closest furniture groups in the $3_{\text{closest}} \in R^3$ vector. Closeness is measured by the $\delta$ function, and furniture groups are stored such that the closest group is the first component of $3_{\text{closest}}$ and the farthest group is the third component.

%\paragraph{Edge and Corner Booleans ($EB$ \& $CB$): }
%$[EB, CB] \in R^2$
% Room position features were introduced in \cite{keshavarzi2020scenegen}. However, for the summary vector, rather than taking a numerical value to encode room position, we choose to use a one-hot encoded vector $[EB, CB] \in R^2$. Given an object to be placed $O_k$ and the room $R$, we utilize the RoomPos function (see section 3.3.1) in order to set $[EB, CB]$:
\begin{equation}
    [EB, CB] = 
    \begin{cases} 
      [1, 0] & \mbox{RoomPos}(O_k, R) = 1; \\ 
      [0, 1] & \mbox{RoomPos}(O_k, R) \geq 2;  \\
      [0, 0] & \text{otherwise}.
   \end{cases}
\end{equation}

% For our purposes, our system trains and tests on rooms with four walls, and thanks to the work of SceneGen for reorienting rooms, walls are consistently placed in the minimum x, minimum y, maximum x, and maximum y planes. 

% Relying on this geometry, we judge nearness in the x direction by calculating whether the distance between a wall and an object bounding box is less than 20\% of the width of the room (width refers to distance between the minimum x wall and the maximum x wall). Nearness in the y direction is defined similarly.

%\paragraph{Average Distance ($AD$): }
%$AD \in R^m$
% The average distance vector $\in R^8$  was introduced by \cite{keshavarzi2020scenegen}, and it contains the average distance per existing furniture group in a room to the proposed placement of our target object $O_k$. In particular, each component in the average distance vector corresponds to a furniture group $g_i$. The value of the $i$th component is equal to AvgDist($O_k$, $g_i$):
\begin{equation}
    AD = [\mbox{AvgDist}(O_k, g_1), ..., \mbox{AvgDist}(O_k, g_m)]
\end{equation}

%\paragraph{Surrounded By ($SB$): }
%$SB \in R^m$
% The surrounded-by vector $\in R^8$ was introduced by \cite{keshavarzi2020scenegen}, and it contains a count per furniture group corresponding to how many objects of the group are surrounding our target object $O_k$. In particular, each component in the surrounded-by vector corresponds to a furniture group $g_i$. The value of the $i$th component is equal to SurrBy$(O_k, g_i)$:
\begin{equation}
    SB = [\mbox{SurrBy}(O_k, g_1), ..., \mbox{SurrBy}(O_k, g_m)]
\end{equation}

%\paragraph{Intersection in XY Plane ($IXY$): } $IXY \in R^{m+1}$, where $m$ components are used for furniture groups and the last component is use for walls $W_R$ of room $R$.

% The intersection-in-XY vector $\in R^9$ contains a count per furniture group corresponding to how many objects of the group have intersecting bounding boxes with the target object bounding box $B(O_k)$. This vector is similar to $AD$ and $SB$ except that it is 9-dimensional because we take into account furniture groups ($g_1$,...,$g_m$) and the room walls $W_R$.
\begin{equation}
    IX = [\mbox{InterXY}(O_k, g_1), ..., \mbox{InterXY}(O_k, W_R)]
\end{equation}

%\paragraph{Supported By and Support To ($SBY$ \& $STO$): }
%$SBY \in R^9$ and $STO \in R^9$. Dimensionality reasoning is the same as $IXY$. % contains a count per furniture group corresponding to how many objects of the group are supporting the target object (hence the target being supported by objects of a furniture group). The support-to vector in $\in R^9$ is very similar to the supported-by vector, except that we now consider whether or not our target object supports other objects. The support spatial relationship is used in multiple instances of scene synthesis literature (\cite{li2019grains}, \cite{zhou2019scenegraphnet}, \cite{keshavarzi2020scenegen}). Similar to InterXY, both vectors are 9-dimensional because we take into account furniture groups ($g_1$,...,$g_m$) and the room walls $W_R$.
\begin{equation}
    SBY = [\mbox{SuppBy}(O_k, g_1), ..., \mbox{SuppBy}(O_k, W_R)]
\end{equation}
\begin{equation}
    STO = [\mbox{SuppTo}(O_k, g_1), ..., \mbox{SuppTo}(O_k, W_R)]
\end{equation}

% \paragraph{Support To ($STO$): }

% The support-to (SuppTo) vector in $R^9$ is very similar to the SuppBy vector, except that we now consider whether or not our target object supports other objects. Therefore, if the distance between the top plane of our target object’s bounding box and the bottom plane of another object’s bounding box is less than 0.05 meters, then we consider this to be a support-to relationship. Again, this vector is 9-dimensional because we are taking into account furniture groups and the walls.

\subsubsection{Initialization} Features vectors associated with nodes in the scene graphs are passed through a 4-layer initialization neural network $\text{INIT}_{g_{O_k}}$, which transforms the dimensionality of the feature vector from 48 dimensions to 100. The resulting node set then becomes $\hat{V}_{r}$ to represent nodes associated with the transformed feature vectors $\hat{V}_{r, \text{feats}}$, and the resulting graph becomes $\hat{\mathcal{G}}_{r} \sim (\hat{V}_{r}, E_{r})$. 
\begin{equation}
\forall r \in \mathcal{R}, \hat{V}_{r, \text{feats}} = \text{INIT}_{g_{O_k}}(V_{r, \text{feats}})
\end{equation}

\subsubsection{Graph Attention} Each scene graph $\mathcal{G}_r$ for a spatial relationship $r$ is fed into its respective attention graph layer $\text{GAT}_{g_{O_k}, r}$. Multi-head attention is suggested to stabilize the learning process, and applying dropout to the attentional coefficients is seen as a highly beneficial regularizer \cite{vaswani2017attention,velivckovic2017graph}. Therefore, for each $\text{GAT}_{g_{O_k}, r}$, we use 10 heads, each with output dimension of 10, and a dropout of 0.8 for each $\text{GAT}_{g_{O_k}, r}$. Concatenating the outputs of each head results in final output vector of dimension 100, and there is a 100-dimensional output vector given to each node in a scene graph. In the message passing context, we consider this vector as the finalized message passed to a node.
\begin{equation}
\forall r \in R, Z_{r} = \text{GAT}_{g_{O_k}, r}(\hat{V}_{r}, E_{r})
\end{equation}
\subsubsection{Projection} After each scene graph is passed through the scene graph attention module, we extract messages passed to the node associated with the furniture $O_k$. We concatenate messages $Z_{r}$ per scene graphs ($n$ total) with the summary vector $X_{O_k}$. We pass the concatenated vector into a 4-layer network $\text{PROJ}_{g_{O_k}}$, which acts as a method to project the concatenated vector into a space such that data points representing plausible placements are clustered together while data points representing unplausible placements are separated from the cluster. Our resulting projected matrix is labelled as $\hat{Y}$.
\begin{equation}
\hat{Y} = \text{PROJ}_{g_{O_k}}([Z_{r_1}, ..., Z_{r_n},  X_{O_k}])
\end{equation}
\subsubsection{Plausibility Assessment}\label{sec:plausAss} Finally, we output a probability of plausible placement $P$ using the reconstruction error produced by an autoencoder $\text{AE}_{g_{O_k}}$. Specifically, the 4-layer encoder of $\text{AE}_{g_{O_k}}$ is given $\hat{Y}$, which converts the input to a coded vector. Then, with the decoder, $\text{AE}_{g_{O_k}}$ will attempt to reconstruct $\hat{Y}$ based off the coded vector. %In previous auto encoder research, 
Autoencoders are shown to carry a built-in anomaly detector because decoders will be able to better reconstruct an input to the encoder if the input has been seen before \cite{zong2018deep}. By training on $\hat{Y}$'s corresponding to real placements of furniture group $l_F$, we allow $\text{AE}_{g_{O_k}}$ to learn real placements as non-anomalies. With this anomaly detection ability of $\text{AE}_{g_{O_k}}$ in mind, suppose we call the output of the decoder as $\hat{Y}^\prime$. We measure the reconstruction error via the mean squared error MSE between $\hat{Y}$ and $\hat{Y}^\prime$, and we use the reconstruction error as the negative log probability of plausibility. Finally, to convert to a valid probability, we use the mean squared error as the power to an exponential function.
\begin{equation}
P = e^{-MSE(\hat{Y}, \hat{Y}^\prime)}
\end{equation}
\subsection{Training}

For our system, we have a model $M_{g_i}$ for each furniture group $g_i \in G$, and each model follows the network architecture described in Section \ref{sec:netarch}. By using a model per $g_i$, each model is trained to specialize in the plausible placement of $g_i$.

We train each model $M_{g_i}$ using two separate training processes. In the first process, we train together IGATP and siamese network projection modules. In the second process, we use the outputs of the first training process as input and train the autoencoder module alone. The following paragraphs detail both training processes.

\begingroup
\setlength{\tabcolsep}{3pt} % Default value: 6pt
\renewcommand{\arraystretch}{1} % Default value: 1

\begin{table}

  \caption{Data augmentation method with the smallest average distance error between ground truth and top-1 (T1) and top-5 (T5) predicted positions for scene augmentation task.}\label{tab:dataaug}
  \centering
  \begin{tabular}{|c|c c||c|c c|}
          \hline 
          \hline
    
  \footnotesize{\textit{Furniture}}  & 
 \tiny{T1} & \tiny{T5}&
 \footnotesize{\textit{Furniture}}&
 \tiny{T1} & \tiny{T5}
 
 \\

    \hline
    \footnotesize{\textit{Bed}}
    & \footnotesize{M3DPIA}
    & \footnotesize{M3DPIA}
    &\footnotesize{\textit{Chair}}
    & \footnotesize{M3DP}
    & \footnotesize{M3DP}
\\

    \footnotesize{\textit{Decor}}
    & \footnotesize{M3DPIA}
    & \footnotesize{M3DPIA}
    &\footnotesize{\textit{Picture}}
    & \footnotesize{M3D}
    & \footnotesize{M3DR4PIA}
\\
    
    \footnotesize{\textit{Sofa}}
    & \footnotesize{M3DPIA}
    & \footnotesize{M3DPIA}
    &\footnotesize{\textit{Storage}}
    & \footnotesize{M3DR4PIA}
    & \footnotesize{M3DR4PIA}
\\

    \footnotesize{\textit{Table}}
    & \footnotesize{M3D}
    & \footnotesize{M3DPIA}
    &\footnotesize{\textit{TV}}
    & \footnotesize{M3DP}
    & \footnotesize{M3DR4PIA}
\\
     \hline
    % \bottomrule
  \end{tabular}
\end{table}

\begin{table*}

  \caption{Average distance error between ground truth and top-1 (T1) and top-5 (T5) predicted positions for scene augmentation task via different models.}\label{tab:results}
  \centering
  \begin{tabular}{|c|c c|c c| c c | c c | c c | c c | c c | c c |c c|}
\hline
\hline
    {\textit{System}}
    & \multicolumn{2}{c}{ \footnotesize{\textit{Bed}}}
      & \multicolumn{2}{c}{ \footnotesize{\textit{Chair}}}
    & \multicolumn{2}{c}{ \footnotesize{\textit{Decor}}}
     & \multicolumn{2}{c}{\footnotesize{\textit{Picture}}}
      & \multicolumn{2}{c}{ \footnotesize{\textit{Sofa}}}
       & \multicolumn{2}{c}{ \footnotesize{\textit{Storage}}}
        & \multicolumn{2}{c}{ \footnotesize{\textit{Table}}}
         & \multicolumn{2}{c}{\ \footnotesize{\textit{TV}}}
          & \multicolumn{2}{c}{ \footnotesize{\textit{Overall}}}\\
          \hline 
          \hline
    
    & 
 \tiny{T1} & \tiny{T5}&
 \tiny{T1} & \tiny{T5}&
 \tiny{T1} & \tiny{T5}&
 \tiny{T1} & \tiny{T5}&
 \tiny{T1} & \tiny{T5}&
 \tiny{T1} & \tiny{T5}&
 \tiny{T1} & \tiny{T5}&
 \tiny{T1} & \tiny{T5}&
 \tiny{T1} & \tiny{T5}
 \\
    \hline
    
    \footnotesize{Siamese         }& \footnotesize{ 1.77       }& \footnotesize{ 1.77       }& \footnotesize{ 3.03        }& \footnotesize{ 2.90         }& \footnotesize{ 2.92        }& \footnotesize{ 2.86        }& \footnotesize{ 3.44         }& \footnotesize{ 3.24         }& \footnotesize{ 3.49        }& \footnotesize{ 3.43       }& \footnotesize{ 2.89         }& \footnotesize{ 2.86         }& \footnotesize{ 2.47        }& \footnotesize{ 2.47        }& \footnotesize{ 3.23       }& \footnotesize{ 3.07      }& \footnotesize{ 2.90 }& \footnotesize{ 2.82}  \\
\footnotesize{GAT+ResNet }& \footnotesize{ 3.64       }& \footnotesize{ 3.64       }& \footnotesize{ 4.89        }& \footnotesize{ 4.89        }& \footnotesize{ 3.75        }& \footnotesize{ 3.75        }& \footnotesize{ 3.53         }& \footnotesize{ 3.53         }& \footnotesize{ 4.43        }& \footnotesize{ 4.43       }& \footnotesize{ 2.48         }& \footnotesize{ 2.48         }& \footnotesize{ 3.86        }& \footnotesize{ 3.86        }& \footnotesize{ 3.53       }& \footnotesize{ 3.53      }& \footnotesize{ 3.72  }& \footnotesize{ 3.72}  \\
\footnotesize{GAT+Siamese     }& \footnotesize{ 2.99       }& \footnotesize{ 2.5        }& \footnotesize{ 2.87        }& \footnotesize{ 2.31        }& \footnotesize{ \textbf{ 2.85}        }& \footnotesize{ 2.45        }& \footnotesize{ \textbf{3.17   }      }& \footnotesize{ 2.08         }& \footnotesize{ 3.05        }& \footnotesize{ 2.55       }& \footnotesize{ 3.12         }& \footnotesize{ 2.63         }& \footnotesize{ 2.91        }& \footnotesize{ 2.35        }& \footnotesize{ 2.62       }& \footnotesize{ 2.37      }& \footnotesize{ 2.98  }& \footnotesize{ 2.35}  \\
\footnotesize{Siamese+KDE     }& \footnotesize{ 2.9        }& \footnotesize{ 2.9        }& \footnotesize{ 3.13        }& \footnotesize{  3.13           }& \footnotesize{ 3.65        }& \footnotesize{ 3.56        }& \footnotesize{ 3.38         }& \footnotesize{ 3.23         }& \footnotesize{ 4.13        }& \footnotesize{ 4.21       }& \footnotesize{ 2.27         }& \footnotesize{ 2.52         }& \footnotesize{ 3.7         }& \footnotesize{ 3.42        }& \footnotesize{ 3.28       }& \footnotesize{ 3.03      }& \footnotesize{ 3.37  }& \footnotesize{ 3.03}  \\
\footnotesize{GAT+Siamese+KDE }& \footnotesize{ 1.75       }& \footnotesize{ 1.14       }& \footnotesize{ 3.25        }& \footnotesize{ 2.31        }& \footnotesize{ 3.14        }& \footnotesize{ 1.75        }& \footnotesize{ 2.99         }& \footnotesize{ 2.24         }& \footnotesize{ \textbf{2.18}        }& \footnotesize{ 1.07       }& \footnotesize{ 2.77         }& \footnotesize{ 1.85         }& \footnotesize{ 2.95        }& \footnotesize{ 2.12        }& \footnotesize{ 2.70        }& \footnotesize{ 1.99      }& \footnotesize{ 2.80  }& \footnotesize{ 1.88} \\

     \hline
\footnotesize{GSACNet (Ours)      }& \footnotesize{ \textbf{1.47}       }& \footnotesize{\textbf{ 1.29  }     }& \footnotesize{\textbf{ 2.45   }     }& \footnotesize{\textbf{ 1.72}        }& \footnotesize{ 3.10        }& \footnotesize{ \textbf{2.06  }      }& \footnotesize{ 3.34         }& \footnotesize{\textbf{ 1.83}         }& \footnotesize{ 2.25        }& \footnotesize{\textbf{ 1.03}       }& \footnotesize{ \textbf{2.48}         }& \footnotesize{ \textbf{1.50   }       }& \footnotesize{ \textbf{2.35}        }& \footnotesize{ \textbf{1.51}        }& \footnotesize{\textbf{ 2.56}       }& \footnotesize{ \textbf{1.06}      }& \footnotesize{ \textbf{2.66}  }& \footnotesize{ \textbf{1.63}}  \\
\footnotesize{SceneGraphNet \cite{zhou2019scenegraphnet}  }& \footnotesize{ 2.77       }& \footnotesize{ 2.42       }& \footnotesize{ 3.56        }& \footnotesize{ 3.11        }& \footnotesize{ 3.51        }& \footnotesize{ 3.02        }& \footnotesize{ 3.64         }& \footnotesize{ 3.21         }& \footnotesize{ 4.59        }& \footnotesize{ 3.83       }& \footnotesize{ 2.88         }& \footnotesize{ 2.51         }& \footnotesize{ 4.04        }& \footnotesize{ 3.58        }& \footnotesize{ 4.46       }& \footnotesize{ 3.98      }& \footnotesize{ 3.61  }& \footnotesize{ 3.15} 
\\

     \hline
    % \bottomrule
  \end{tabular}
\end{table*}

\endgroup

\subsubsection{Siamese Learning} In this training process, we consider the initialization, scene graph extraction, and project modules as one large siamese network IGATP. %Training siamese networks offer two advantages: (1) they don't require an enormous amount of data and (2) they learn to cluster similar labeled data points and separate differently labeled data points via a learned projection into another feature space. 
In our case, labels are binary where 1 means a plausible placement for $g_i$ and 0 means otherwise. 

$\forall i \in \{1,2\}$, unprocessed input $D_i$ contains a room $R_i$, a furniture to be placed $O_i$ of furniture group $g_{O_i}$, and placement center $\dot{O_i}_{xy}$, and $L_i \in \{0,1\}$ describes whether or not $O_i$ centered at $\dot{O_i}_{xy}$ in $R_i$ is plausible. We train this siamese network by giving pairs of unprocessed input. Suppose we consider the first unprocessed data point as $D_1$ and the second as $D_2$, along with their labels $L_1$ and $L_2$. Scene graphs and summary vectors are extracted from $R_i$ from the perspective of $O_i$, and as described in Section \ref{sec:netarch}, IGATP takes these scene representations as input and outputs a vector $\hat{Y}_{i}$. Therefore, the output associated with $D_1$ and $D_2$ would be $\hat{Y}_{1}$ and $\hat{Y}_{2}$, respectively.

Now that we have $\hat{Y}_{1}$ and $\hat{Y}_{2}$, we calculate the max margin contrastive loss $\mathcal{L}$ between these two outputs.

$$ \mathcal{L}(\hat{Y}_{1}, \hat{Y}_{2}) = \begin{cases}
        \|\hat{Y}_{1}- \hat{Y}_{2}\|^2_2 & L_1 = L_2 \\
        \text{max}(0, m-\|\hat{Y}_{1}- \hat{Y}_{F_2}\|^2_2) & L_1 \neq L_2 \\
    \end{cases}$$

$m > 0$ is the margin parameter for the contrastive loss function, and it acts as a lower bound on the distance between a pair of data points with different labels (i.e. $L_1 \neq L_2$).% We choose $m=15$.
After calculating the contrastive loss, we backpropogate the loss to update weights across the siamese network.% In regard to training implementation details, we train this network for 100 epochs with a batch size of 100 pairs.  Additionally, summary vectors are standardized, and we use an Adam optimizer with a learning rate of 0.005 and an L2 regularization weight of 1.

\subsubsection{Autoencoder Training} In this training pipeline, we use $\hat{Y}_{i}$ from the trained siamese network as input to train the autoencoder $AE_{g_{O_i}}$. All $\hat{Y}_{i}$ correspond to $L_i=1$ because we want the autoencoder to familiarize itself with plausible placements of furniture group $g_{O_i}$. As a result, the reconstruction error of plausible ($L_i=1$) $\hat{Y}_{i}$ will be low, while the reconstruction error of implausible ($L_i=0$) $\hat{Y}_{i}$ will be high. From the perspective of anomaly detection, input vectors associated with implausible layouts will be regarded as anomalies.

As described in Section \ref{sec:plausAss}, the encoder of $AE_{g_{O_i}}$ takes in $\hat{Y}_{i}$ as input and transforms the input vector into another vector of smaller dimensionality to create a bottleneck effect. On the other end, the decoder is forced to use the smaller vector to reconstruct the input. We use the mean squared error $MSE$ between the input and the output as the reconstruction error $RE$.
$$RE = MSE(\hat{Y}_{i}, \hat{Y}_{i}^\prime)$$
Finally, we backprogate $RE$ to all the layers of $AE_{g_{O_i}}$. % In regard to implementation details, we train this auto encoder using 100 epochs and a batch size of 100 data points. We use an Adam optimizer with a learning rate of 0.005 with a L2 regularization weight of 0.

\subsection{Data Preparation}
\subsubsection{Dataset}
We use Matterport3D (M3D) {\cite{Chang2018}} which consists of various building types with diverse architecture styles, including numerous spatial functionalities and furniture layouts. Annotations of building elements and furniture are provided with surface reconstruction as well as 2D and 3D semantic segmentation. For this study, we reduce the categories of object types considered for building our model and placing new objects. We group the objects into 8 coarse categories: $G =$ \{Bed, Chair,  Decor, Picture, Sofa, Storage, Table, TV\}. For room types, we consider the set $\{$Library,  Living Room, Meeting Room, TV Room, Bedroom, Recreation Room, Office, Dining Room, Family Room, Kitchen, Lounge$\}$ to avoid overly specialized rooms such as balconies, garages and stairs. We also filter rooms which hold more than 95\% unoccupied areas to avoid unusual empty rooms that come without any spatial arrangements.% After the data reduction, a total of 1,326 rooms and 7,017 objects are in our training and validation sets.

\subsubsection{Parametric Data Augmentation}
We use a modified version of the scene augmentation method introduced in \cite{keshavarzi2020genscan}.
%We calculate the center coordinates of each bounding box assigned to objects and perform a closest point search with a parametric wall system to find the closest wall. We then transform each object with the two-dimensional position translation vector of the corresponding closest wall, with a non-linear factor of its distance to the wall so that an object closer to the wall would have a much similar transformation function to the wall itself, than an object located in the middle of the room. This would allow furniture to move close and far in relation to each other, instead of moving in a similar direction altogether. During the transformation, the proportional distance of the object with the adjacent walls to the corresponding wall is maintained.
Furthermore, we run two sets of area checks: (a) to check whether the area of the open space of the room is not larger than a certain percentage of the overall area of the room. This would disqualify the augmentations which result in overly large rooms with extensive open space. Next, (b) we check whether the intersection of two non-colliding objects is not larger than a percentage of the smaller object. Finally, we run another round of data augmentation by removing $n$ smallest objects for the generated scenes. Figure \ref{fig:augmentation-example} illustrates an example of the parametric data augmentation for a given room.

%\begin{figure}
%  \includegraphics[width=1\columnwidth]{figures/15_bedroom_JeFG25nYj2p.jpg}
%  \caption{Example of parametric data augmentation.}~\label{fig:augmentation-example}
%\end{figure}

\section{Experiments}
To evaluate our prediction system, we run ablation studies, examining how the presence or absence of particular features affects our prediction results. We use a subset of our dataset which include 200 room with a 4-fold cross validation method and a 80/ 20 split between the training and validation set. In these studies, we remove each object in the validation set, one at a time, and use our model to predict where the removed object should be positioned. We compute the distance between the original object location and our system's top prediction. We also compute the the smallest distance to the top 5 predictions to address the multi-model property of objects which can be placed in several valid locations.

\subsection{Data Augmentation}
In the data augmentation experiments, we prepare four datasets. The original Matterport3D dataset (M3D), the M3D dataset with Parametric Data Augmentation (M3DP),  the M3DP dataset with area and intersection checks (M3DPIA), and the M3DPIA dataset + 4 smallest items iteratively removed from a room to create 4 new rooms (M3DR4PIA). By conducting the object removal experiment, we aim to find which of the mentioned datasets achieve lower distances errors for each object category via the GSACNet model. As shown in Table \ref{tab:dataaug}, we find that the data augmentation proposed in this study effectively improves the scene augmentation workflow.

\subsection{Comparative Studies}
We compare the performance of our system with alternative learning models and also SceneGraphNet \cite{zhou2019scenegraphnet}.
The results can be seen in Table \ref{tab:results} in which GSACNet outperforms alternative learning models and  \cite{zhou2019scenegraphnet} in nearly all categories. Details of the implementation of various 
models can be found in the supplementary material.

\section{Conclusion}
The network that we presented in this paper takes a novel approach to contextual scene augmentation through a Graph Attention and Siamese network architecture followed by a autoencoder network and its implementation of parametric data augmentation of a 3D space with objects. We find that utilizing such model improves the ability to to augment virtual objects in plausible placements in a scene despite a small set of training data. By training on parametrically augmented version of the Matterport3D dataset, we show our network architecture outperforms state-of-the-art scene synthesis networks such as  \cite{zhou2019scenegraphnet}.
% while providing comparable results with \cite{keshavarzi2020scenegen}.
Our work comes with a number of limitations. First, our current system does not conduct pose estimation for the augmented objects. Moreover, in multi-object placement scenarios, the resulting predictions are highly dependent on the order in which object are to be placed. Such approach does not consider all possible combinations of the possible arrangements. Future work can comprise of incorporating  floorplanning methodologies with the current sampling mechanism allowing a robust search in the solution space, while addressing combinatorial arrangement.

{\small
\bibliographystyle{ieee_fullname}
\bibliography{egbib}
}

\appendix
\section{Implementation Details}
\label{sec:implementation}

We rely on the implementation of graph attention layers and dynamic edge convolution layers from Pytorch Geometric \cite{fey2019fast}.
%In order to parellize training on GPUs, we utilize Ray \cite{moritz2018ray}. We use 2 Nvidia 3090 GPUs to conduct training and scene synthesis.
For the  \textit{RoomPosition} feature, $\rho$ is set to 20\% of the width and length of $R$. In the \textit{support} feature, the threshold is equal to 0.05 meters. For siamese learning, we choose $m=15$ margin parameter for the contrastive loss function. In regard to training implementation details, we train this network for 100 epochs with a batch size of 100 pairs.  Additionally, summary vectors are standardized, and we use an Adam optimizer with a learning rate of 0.005 and an L2 regularization weight of 1. For the autoencoder training, we train using 100 epochs and a batch size of 100 data points. We use an Adam optimizer with a learning rate of 0.005 with a L2 regularization weight of 0.

\subsection{Parametric Data Augmentation}

Below we provide additional details of the modified version of the scene augmentation method introduced in \cite{keshavarzi2020genscan}, and our data filtering techniques for the parametric data augmentation.

We calculate the center coordinates of each bounding box assigned to objects and perform a closest point search with a parametric wall system to find the closest wall. We then transform each object with the two-dimensional position translation vector of the corresponding closest wall, with a non-linear factor of its distance to the wall so that an object closer to the wall would have a much similar transformation function to the wall itself, than an object located in the middle of the room. This would allow furniture to move close and far in relation to each other, instead of moving in a similar direction altogether. During the transformation, the proportional distance of the object with the adjacent walls to the corresponding wall is maintained. We applied two filters.

We filter rooms which hold more than 95\% unoccupied areas to avoid unusual empty rooms that come without any spatial arrangements. We also filter items by establishing a list of items that could overlap with each other and items whose volumes overlap more than 40\% of the smallest item are filtered out.  In addition, we remove \textit{n} of the smallest items by volume in a iterative manner. After the removal of an item, the updated room is saved. For example, if \textit{n}=2, two new rooms would be generated. In the first room only the smallest items is removed and in the second room the two smallest items are removed. We use \textit{n}=4 in our experiments, in which four new rooms are be generated for each room. We run a further check to make sure that all rooms have at least one items remaining after iterative object removal. Using the techniques mentioned above, we created four main datasets. The item counts for and number of rooms can be seen in  Table \ref{tab:objectCountSmall}

\begin{table}

  \caption{Total object and room counts in different data augmentation techniques. M3D is sampled from the the original Matterport3D \cite{Chang2018} dataset. M3DP is the M3D dataset which has gone through parametric data augmentation. M3DPIA is M3DP with the two filtering techniques applied. M3DR4PIA is M3DPIA with iterative smallest item removal of n = 4. }\label{tab:objectCountSmall}
  \centering
  \begin{tabular}{|c |c c c c|}
          \hline 
    \hline
    \footnotesize{\textbf{\textit{Object Type}}}  & 
        M3D & M3DP & M3DPIA & M3DR4PIA  \\
    \hline

    \footnotesize{\textit{Bed}}
    & \footnotesize{155}
    & \footnotesize{3100}
    &\footnotesize{720}
    & \footnotesize{4313}
  
\\
    
    \footnotesize{\textit{Chair}}
    & \footnotesize{118}
    & \footnotesize{2360}
    &\footnotesize{674}
    & \footnotesize{3280}

\\

    \footnotesize{\textit{Decor}}
    & \footnotesize{267}
    & \footnotesize{5340}
    &\footnotesize{1473}
    & \footnotesize{5748}

\\

\footnotesize{\textit{Picture}}
    & \footnotesize{367}
    & \footnotesize{7340}
    &\footnotesize{1192}
    & \footnotesize{4041}

\\

\footnotesize{\textit{Sofa}}
    & \footnotesize{109}
    & \footnotesize{2180}
    &\footnotesize{446}
    & \footnotesize{2704}

\\

\footnotesize{\textit{Storage}}
    & \footnotesize{125}
    & \footnotesize{2500}
    &\footnotesize{436}
    & \footnotesize{2601}

\\

\footnotesize{\textit{Table}}
    & \footnotesize{260}
    & \footnotesize{5200}
    &\footnotesize{1140}
    & \footnotesize{6251}

\\

\footnotesize{\textit{TV}}
    & \footnotesize{45}
    & \footnotesize{900}
    &\footnotesize{158}
    & \footnotesize{604}

\\
     \hline
   
   \footnotesize{\textbf{\textit{Rooms}}}
    & \footnotesize{179}
    & \footnotesize{3580}
    &\footnotesize{810}
    & \footnotesize{4605} 
 \\   
     \hline
     
    % \bottomrule
  \end{tabular}
\end{table}

\begin{figure*}
  \includegraphics[width=2\columnwidth]{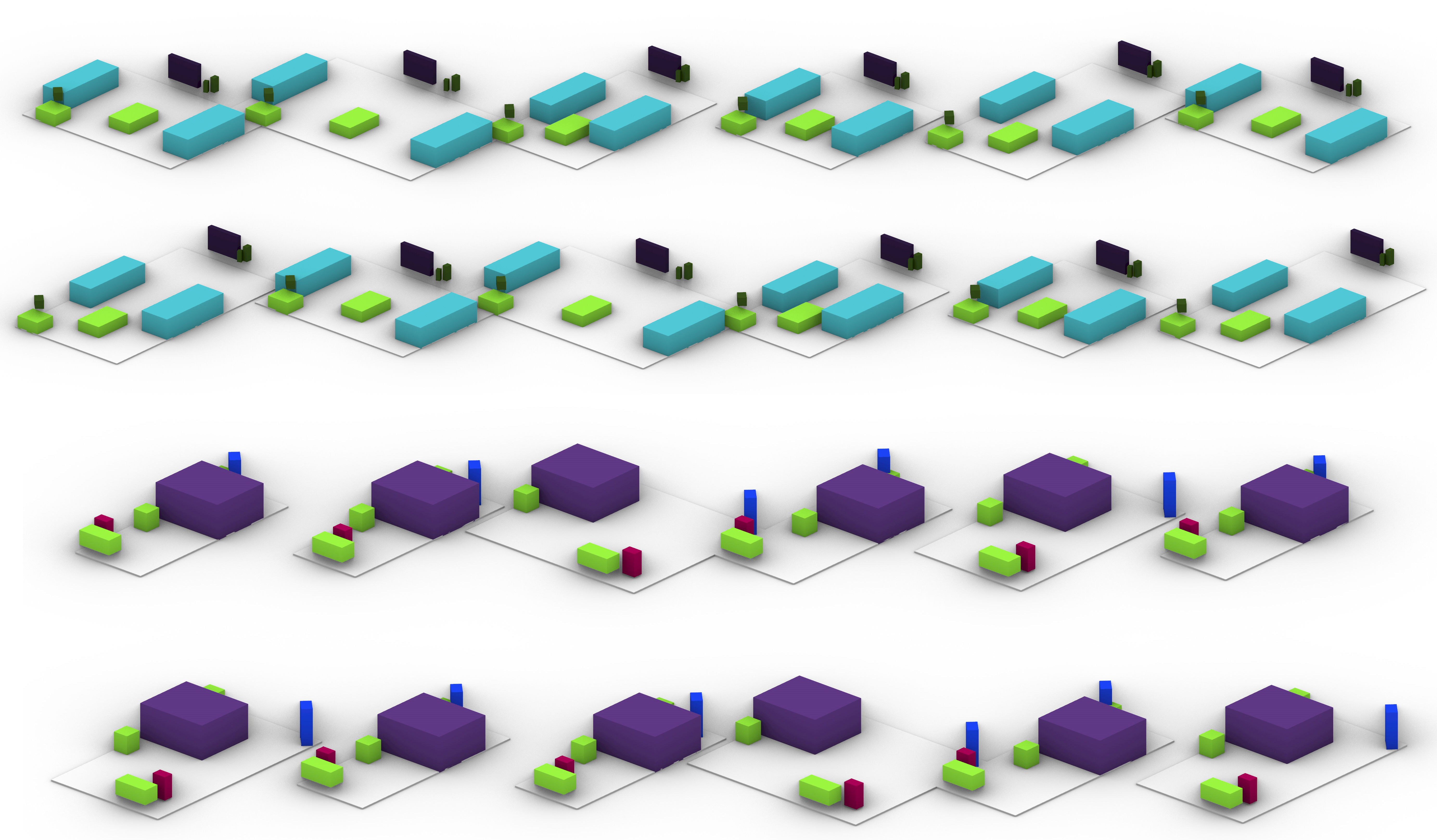}
  \caption{Example of parametric data augmentation.}~\label{fig:augmentation-example}
\end{figure*}

\subsection{Ablation Models}
Description of the ablation models utilized in Table 2 of the original paper can be found below. As a point of clarification, there is a model per furniture group $g_i$, and each model assess plausibility of placement for $g_i$ during the scene synthesis workflow.

\subsubsection{Siamese}\label{sec:siamese}

The siamese model $S$ is a 3 layer fully connected neural network with ReLU activations, followed by an additional linear activation layer. The model takes  a summary vector $X_{O_k}$ as input. We train $S$ per furniture group $g_i$ using pairs of data points whose outputs will be fed into the contrastive loss function, similar to the training pipeline described in section 3.4.1 of the original paper without any scene graphs or graph neural network layers. In order to represent this granularity, we define each model per furniture group as $S_{g_i}$. After training, to assess plausibility at scene synthesis time, we first create clusters per furniture group in the siamese learned space. In particular, for each furniture group $g_i$, we extract a summary vector $X_{O_k}$ for each object $O_k \in g_i$ and get the output of the model $S_{g_i}(X_k)$. Then, we calculate the mean of the outputs for furniture group $g_i$:
\begin{equation}
    \mu_{S_{g_i}} = \frac{1}{|g_i|}\sum_{O_k \in g_i} S_{g_i}(X_k).
\end{equation}

At scene synthesis time, for an object $O_j \in g_i$ proposed to be placed at $\dot{O_j}(x,y)$, $\|S_{g_i}(O_j) - \mu_{S_{g_i}}\|_2 = d_{O_j, \mu_{S_{g_i}}}$ is calculated. Then, the probability of plausible placement is calculated as follows:
\begin{equation}
    P = e^{-d_{O_j, \mu_{SIAM_{g_i}}}}.
\end{equation}

\subsubsection{GAT + ResNet}\label{sec:gatresnet}

The GAT + ResNet model utilizes graph attention layers to conduct inference on scene graphs and a ResNet34 model to extract features from a scene image \cite{he2015deep}. Inspired by the work of \cite{wang2019planit}, the scene image is defined as a $R^{100\times100\times15}$ tensor. It serves as an enhanced top down view in the sense that the channels encode information about objects in the scene. There exists padding (0.5 meters) between the boundary of the image and the walls of the room. Additionally, 100 pixels along of the length of the image represent 100 points sampled uniformly in space along the length of the scene plus padding; this is the same for width. The first 11 channels are occupancy maps per furniture group plus walls and floors, the 12th channel is an occupancy map for XY intersections, the 13th channel is an occupancy map for any furniture plus walls, the 14th channel is a height map where the $ij$ element represents the height corresponding to the $ij$ sample point of the room (recall that the image is a sampling of the scene).

The ResNet34 model extracts a 600 dimensional feature vector from the scene image, and the scene graph extraction system extracts a 600 dimensional feature vector from the scene graphs provided. These two 600 dimensional vectors are concatenated with a 48 dimensional summary vector, and this large vector is passed through a fully connected neural network. This neural network is 4 layers deep with ReLU activations, and there is one more layer at the end which has a sigmoid activation. The output of this network gives the probability of plausibility.

\subsubsection{GAT + Siamese}\label{sec:gatsiamese}

The GAT + Siamese model utilizes graph attention layers with one dynamic ddge convolution layer for the co-occurence scene graph. The scene graph extraction system extracts a 600 dimensional feature vector, and it is concatenated with a summary vector extracted from a scene. Then, this larger vector is passed through a fully connected neural network that is 4 layers deep with ReLU activations, and there is one more layer at the end with a linear activation. The model as a whole is trained using pairs of data points and the contrastive loss function; the training pipeline is the same as described in section \ref{sec:siamese} of the supplementary material. Additionally, the plausibility assessment strategy is that same what was described in section \ref{sec:siamese} of the supplementary material.

\subsubsection{Siamese + KDE}\label{siamesekde}

This model utilizes the same architecture as mentioned in section 3.1.1 of the supplementary material. However, the plausibility assessment is done using a Kernel Density Estimator (KDE). In particular, there is a KDE associated with each $S_{g_i}$ that is fitted on the outputs $S_{g_i}(X_k), \forall O_k \in g_i$. By fitting the KDE, we are able to calculate the probability $P(S_{g_i}(O_a) | \mathcal{D}_{g_i})$ where $\mathcal{D}_{g_i} = \{S_{g_i}(O_b)| \forall O_b \in g_i \wedge O_b \in \mathcal{D}\}$, $\mathcal{D}$ is our available training set, and $O_a \in g_i$ but may not be in $\mathcal{D}$. Then, at scene synthesis time, for an object $O_j \in g_i$ proposed to be placed at $\dot{O_j}(x,y)$, suppose the output associated with the proposed placement of $O_j$ is $\hat{Y} = S_{g_i}(X_j)$. Then, using $\mathrm{KDE}_{g_i}$, the plausibility of placement is calculated as follows:

\begin{equation}
    P = P(\hat{Y} | \mathcal{D}_{g_i}).
\end{equation}

\subsubsection{GAT + Siamese + KDE}

This model utilizes the same architecture as mentioned in section \ref{sec:gatsiamese} of the supplementary material with the plausibility assessment conducted via a KDE. This strategy is the same as what is described in section \ref{sec:gatsiamese} of the supplementary material.

%\subsubsection{Comparative Studies}
%--- Comment: This is more the explanation of the experiment and not how SceneGraphNet works. 
%We compare our results with results from SceneGraphNet \cite{zhou2019scenegraphnet}. SceneGraphNet takes in a point and uses a neural message passing approach to augment an input 3D indoor scene with new objects matching their surroundings. To compare with SceneGraphNet, we remove an object from a room and compare recommended placement from SceneGraphNet with the ground truth. To do this, we set up a mesh grid representing all possible (x,y) coordinates to evaluate the entire floor space with a [15x15] grid. Then we feed each position value as an input to predict the probability distribution for all categories at that position. This results in a [1 x number of object categories] probability array. For us the shape is [1 x 8]. The last step repeats for all grid positions and after the step we have a [15 x 15 x 8] matrix. To visualize the distribution map for the desired object category index, we used the matrix from the previous step and queried at [: ,: , object category index]. This is used to get the top-K locations to place the desired object. The model is trained and evaluated using the code publicly available for SceneGraphNet.

\end{document}